\documentclass[letterpaper]{article} 
\usepackage{aaai25}  
\usepackage{times}  
\usepackage{helvet}  
\usepackage{courier}  
\usepackage[hyphens]{url}  
\usepackage{graphicx} 
\usepackage{natbib}  
\usepackage{caption} 
\usepackage{algorithm}
\usepackage{algorithmic}
\usepackage{array}
\usepackage[caption=false,font=normalsize,labelfont=sf,textfont=sf]{subfig}
\usepackage{textcomp}
\usepackage{url}
\usepackage{verbatim}
\usepackage{graphicx}
\usepackage{cite}
\usepackage{amsmath}
\usepackage{amssymb}
\usepackage{booktabs}

\usepackage{enumitem}
\usepackage{times}
\usepackage{epsfig}
\usepackage{amsfonts}
\usepackage{bm}
\usepackage{multirow}
\usepackage{makecell}
\usepackage{bbding}
\usepackage[table]{xcolor}
\definecolor{Klein_Blue}{rgb}{0.0, 0.129, 0.6}
\definecolor{cGreen}{RGB}{100,180,100}
\definecolor{cRed}{RGB}{220,50,0}

\usepackage{newfloat}
\usepackage{listings}
\DeclareCaptionStyle{ruled}{labelfont=normalfont,labelsep=colon,strut=off} 
\floatstyle{ruled}
\newfloat{listing}{tb}{lst}{}
\floatname{listing}{Listing}
\title{Exploring Enhanced Contextual Information for Video-Level Object Tracking}
\author{
    Ben Kang\textsuperscript{\rm 1,}\textsuperscript{\rm 2},
    Xin Chen\textsuperscript{\rm 1},
    Simiao Lai\textsuperscript{\rm 1},
    Yang Liu\textsuperscript{\rm 1}\thanks{Corresponding author.},
    Yi Liu\textsuperscript{\rm 3},
    Dong Wang\textsuperscript{\rm 1,}\textsuperscript{\rm 2}
}

\affiliations{
    \textsuperscript{\rm 1}	Dalian University of Technology,\textsuperscript{\rm 2}	Ningbo Institute of Dalian University of Technology,\textsuperscript{\rm 3}	Baidu Inc.\\
    kangben@mail.dlut.edu.cn,
    chenxin3131@mail.dlut.edu.cn,
    laisimiao@mail.dlut.edu.cn,
    ly@dlut.edu.cn,
    liuyi22@baidu.com,
    wdice@dlut.edu.cn
}

\usepackage{bibentry}

\begin{document}

\maketitle

\begin{abstract}
Contextual information at the video level has become increasingly crucial for visual object tracking. However, existing methods typically use only a few tokens to convey this information, which can lead to information loss and limit their ability to fully capture the context. To address this issue, we propose a new video-level visual object tracking framework called \textbf{MCITrack}.  It leverages Mamba's hidden states to continuously record and transmit extensive contextual information throughout the video stream, resulting in more robust object tracking. The core component of MCITrack is the Contextual Information Fusion module, which consists of the mamba layer and the cross-attention layer. The mamba layer stores historical contextual information, while the cross-attention layer integrates this information into the current visual features of each backbone block. This module enhances the model's ability to capture and utilize contextual information at multiple levels through deep integration with the backbone. Experiments demonstrate that MCITrack achieves competitive performance across numerous benchmarks. For instance, it gets 76.6\% AUC on LaSOT and 80.0\% AO on GOT-10k, establishing a new state-of-the-art performance. 
Code and models are available at \url{https://github.com/kangben258/MCITrack}.
 

\end{abstract}

%

\section{Introduction}
Visual object tracking is a crucial task in computer vision, aiming to identify an object in the initial frame of a video and predict its location in subsequent frames. 
Existing tracking methods ~\cite{ATOM,SiamBAN, simtrack,vipt,sbt} typically employ image-level matching techniques, using template and search region images for feature matching and then making predictions based on the matching results. These methods primarily focus on the initial appearance of the target and do not utilize the contextual information available in the video sequence.

\begin{figure}[t]
\begin{center}
\includegraphics[width=0.9\linewidth]{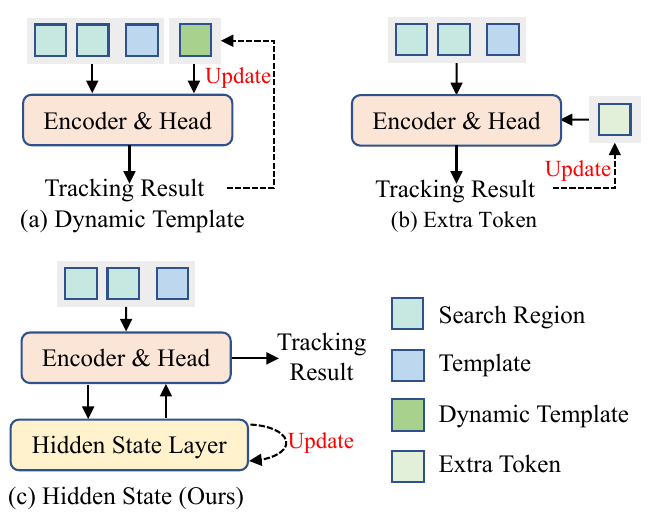}
\end{center}
   \caption{
   Comparison of different contextual information propagation methods. (a) Using dynamic templates to record the object's shape. (b) Using extra tokens to transmit contextual information. (c) Using hidden state layers to transmit contextual information.} 
\label{fig:pipeline}
\end{figure}

To improve performance, some methods~\cite{Stark,ToMP,transt_journal} use dynamic templates to record the appearance information of objects. As shown in Figure~\ref{fig:pipeline} (a), these methods obtain more appearance information and enhance the model's robustness by updating the dynamic template. While these methods have achieved some success, they still primarily focus on the appearance information of the target and do not fully utilize the  contextual information in the video sequence.

Recently, some works~\cite{stid,odtrack,artrackv2,evptrack,aqatrack} have begun to explore video-level object tracking. Unlike previous methods focusing solely on the target's appearance, these works utilize more contextual information to enhance model performance. 
As shown in Figure~\ref{fig:pipeline} (b), these methods typically use a small number of extra tokens to capture and transmit contextual information, continuously updating as the video sequence progresses. 
Although these methods show superior performance compared to earlier approaches, the limited number of tokens can only transmit a limited amount of contextual information, leading to information loss.  Therefore, efficiently transmitting richer contextual information remains an urgent challenge.

Temporal models like LSTM~\cite{lstm} and state space models~\cite{ssm} are adept at preserving  crucial information in their hidden states, making them suitable for recording important contextual information in video sequences.
To address the issues mentioned above, we propose a new framework for transmitting contextual information in video-level object tracking. As shown in Figure~\ref{fig:pipeline} (c), we use hidden state layers to store and transmit richer contextual information.  Specifically, we introduce MCITrack, which leverages hidden states from Mamba~\cite{mamba} to record and transmit contextual information. Compared to using a small number of additional tokens, MCITrack can record and transmit more implicit contextual information, thereby enhancing the model's performance. 

The core module of MCITrack is the Contextual Information Fusion (CIF) module, which consists of mamba and cross-attention layers. The CIF module inputs multi-level features from the backbone blocks of the current frame into the mamba layers, where  hidden states record crucial information. As the next frame is processed,  the cross-attention layers integrate this recorded information into each backbone block. This deep integration allows the CIF module to effectively extract and utilize contextual information at multiple levels, aiding the model in making accurate predictions. Throughout the video, MCITrack continuously updates the hidden states, ensuring the effective transmission of contextual information as the sequence progresses.

Extensive experiments demonstrate the effectiveness of our contextual information transmission framework. MCITrack achieves state-of-the-art performance on multiple datasets. For example, compared to the recent state-of-the-art  model ODTrack-B384~\cite{odtrack}, MCITrack-B224 improves the AUC on LaSOT~\cite{lasot} by 2.1\% (75.3\% $vs.$ 73.2\%) with lower computational cost (38G $vs.$ 73G), fewer parameters (88M $vs.$ 92M), and a lower resolution (256 $vs.$ 384). Even when compared to the larger ODTrack-L384, MCITrack-B224 achieves a 1.3\% higher AUC on LaSOT. Notably, our largest model, MCITrack-L384, reaches an unprecedented AUC of 76.6\% on LaSOT. These results highlight the effectiveness of our framework in transmitting critical contextual information, enhancing prediction accuracy beyond previous methods. The contributions of this paper are summarized as follows:
\begin{itemize}[leftmargin=0.5cm]
    \item{We introduce a new method for transmitting contextual information in video-level object tracking. Compared with previous methods, our approach effectively transmits richer and more crucial contextual information.}
    \item{Based on the proposed framework, we develop a new family of video-level object tracking models named MCITrack. Experiments demonstrate the effectiveness of this framework, with MCITrack achieving state-of-the-art performance across multiple datasets.}
\end{itemize}

\section{Related Work}
\textbf{Visual Tracking.} Current trackers  generally fall into two categories: image-level trackers that rely solely on appearance information and video-level trackers that incorporate more contextual information. Most trackers~\cite{SINT,SiameseFC,SiameseRPN, SiamRPNplusplus,Deeper-wider-SiamRPN, SiamFC++,SiamRCNN, AiATrack,transinmo}  belong to the first category. These trackers use only the initial appearance information of the target to match with the current frame and then predict the target's location based on the matching results through a tracking head. Such trackers often struggle with challenges like object deformation and blurriness.
Some trackers~\cite{stmtrack, TMT, Stark, keeptrack, seqtrack} introduce additional templates, continuously updating them to allow the model to extract more appearance information about the target to address these challenges. Despite these trackers having achieved some success, they still rely solely on appearance for image-level matching to locate the target. They do not utilize the contextual information in the video.

Recently, many researchers have begun exploring video-level object tracking to better utilize contextual information for assisting tracking. TCTrack~\cite{tcttrack} aggregates contextual information through online temporal adaptive convolution and temporal adaptive feature map refinement. VideoTrack~\cite{videotrack} employs a video transformer backbone to integrate contextual information. Besides these methods, scholars have also started exploring the use of additional tokens for this purpose. ODTrack~\cite{odtrack} employs a token sequence propagation paradigm to densely associate contextual relationships across video frames. ARTrack~\cite{artrack} and ARTrackV2~\cite{artrackv2} encode previously predicted object coordinate sequences into tokens to assist the current prediction. AQATrack~\cite{aqatrack} uses learnable autoregressive queries to capture contextual information. EVPTrack~\cite{evptrack} utilizes spatio-temporal markers to propagate information across consecutive frames.

 Compared with previous methods, video-level object tracking models incorporating contextual information have significantly improved tracking performance. However, these models typically rely on a limited number of additional tokens to capture contextual information, which restricts the ability to capture comprehensive contextual information, leading to information loss. In contrast, our MCITrack efficiently captures richer and more critical contextual information, enhancing the model's performance.

\begin{figure*}[t]
\begin{center}
\includegraphics[width=1\linewidth]{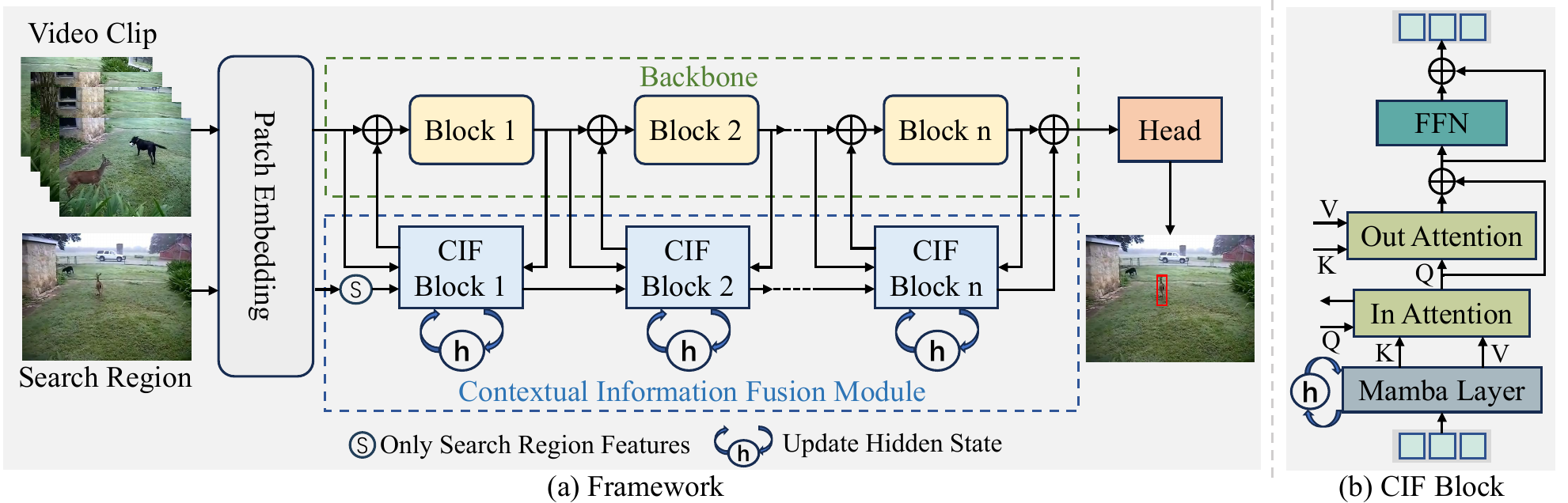}
\end{center}
   \caption{(a) Framework of the proposed MCITrack. The key components include the backbone for visual feature extraction, the contextual information fusion module  for recording and transmitting contextual information, and the prediction head. (b) Architecture of the proposed Contextual Information Fusion(CIF) block.} 
\label{fig:framework}
\end{figure*}

\noindent\textbf{State Space Model.} The State Space Model (SSM) is proposed for handling sequence tasks in NLP~\cite{ssm,ssm1} and has shown potential in addressing long-range dependency problems. Mamba~\cite{mamba} addresses SSM's inability to perform content-based reasoning by making SSM parameters functions of the input. Mamba has demonstrated the powerful capability of SSM in handling long sequences with higher throughput compared to Transformers.  
Subsequently, it has also become popular in the computer vision field. In image classification, ViM~\cite{vim} and VMamba~\cite{vmamba} have achieved performance comparable to ViT~\cite{ViT} with fewer parameters. In the medical image domain, SSM has found widespread applications. For instance, Mamba-UNet~\cite{mamba-unet} incorporates the Mamba structure into the UNet~\cite{unet} model for medical image segmentation. In this work, we explore how to efficiently capture and transmit contextual information in video sequences using the hidden states in Mamba.

\section{MCITrack}
This section presents MCITrack in detail. First, we provide an overview of MCITrack. Next, we describe the model architecture, including the backbone and the Contextual Information Fusion (CIF) module. Finally, we introduce the prediction head and training objective function.
\subsection{Overview}
The overall architecture of MCITrack is shown in Figure~\ref{fig:framework} (a). It consists of three parts: a backbone for visual feature extraction, a Contextual Information Fusion (CIF) module for storing and transmitting contextual information, and a prediction head for making predictions. 
MCITrack takes a video clip and a search region as input.  First, the video clip and search region are divided into patches through patch embedding, and then these patches are concatenated along the spatial dimension and fed into the backbone for feature extraction.
The backbone is composed of $N$ blocks, with each block paired with a corresponding CIF block. The CIF blocks integrate the historical contextual information into their associated backbone blocks, enhancing the accuracy of visual feature extraction based on the historical contextual information. Simultaneously, the CIF blocks update the hidden states based on the current backbone output. Finally, the backbone, guided by contextual information, outputs more precise visual features, which are then passed to the prediction head to obtain the tracking results.

\subsection{Model Architecture}
\subsubsection{Backbone.}  
We use Fast-iTPN~\cite{fastitpn} as our backbone. Similar to ViT~\cite{ViT}, Fast-iTPN consists of Transformer layers. However, compared to ViT, Fast-iTPN is "narrower and deeper", featuring a smaller hidden dimension and more network layers. This deeper structure enables the CIF module to capture a broader range of information, including both shallow-level details and deeper semantic information. Our model takes a video clip $\mathbf{V} \in {\mathbb{R}}^{N \times 3 \times {H_{z}} \times {W_{z}}}$ and a search region  $\mathbf{X} \in {\mathbb{R}}^{3 \times {H_{x}} \times {W_{x}}}$ as inputs.  First, both $\mathbf{V}$ and $\mathbf{X}$ are downsampled using convolutional layers with a stride of 4. The downsampled $\mathbf{V}$ and $\mathbf{X}$ are processed through MLP layers and two convolutional merging layers. This process segments $\mathbf{V}$ and $\mathbf{X}$ into patches, resulting in $\mathbf{F_{v}} \in {\mathbb{R}}^{N \times C \times {\frac{H_{z}}{16}} \times {\frac{W_{z}}{16}}}$ and $\mathbf{F_{x}} \in {\mathbb{R}}^{C \times {\frac{H_{x}}{16}} \times {\frac{W_{x}}{16}}}$. These patches are expanded and concatenated along the spatial dimension to form $\mathbf{F_{vx}} \in {\mathbb{R}}^{C \times L}$ ($L = N \times {\frac{H_{z}}{16}} \times {\frac{W_{z}}{16}} + {\frac{H_{x}}{16}} \times {\frac{W_{x}}{16}}$), which are then input into the Transformer layers for visual feature extraction. To enable the CIF module to capture more information at multiple levels, we divide the backbone's Transformer layers into $N$ ($N$ is 4 by default) blocks, each paired with a corresponding CIF block. Before the features are input into each block, the historical contextual information from the CIF block is integrated,  enhancing the visual feature extraction process. This integration of contextual information aids the backbone in extracting more accurate features for subsequent predictions by the prediction head.


\subsubsection{Contextual Information Fusion Module.} 
The Contextual Information Fusion (CIF) module is the core component of MCITrack. It stores and integrates contextual information into the backbone to enhance visual feature extraction. The CIF module consists of $N$ CIF blocks, each corresponding to a backbone block. Each CIF block integrates contextual information into the backbone features before the corresponding backbone block performs feature extraction, and updates the hidden states based on the current frame's features. This deep integration enables the CIF module to extract information at multiple levels, providing richer implicit contextual information to improve the model's predictions. The structure of the CIF block is shown in Figure~\ref{fig:framework} (b) and primarily includes a mamba layer and two cross-attention layers. The mamba layer is responsible for storing contextual information. The in-attention layer integrates the contextual information from the CIF block into the backbone, while the out-attention layer extracts current frame information from the backbone's features into the CIF block. 
Specifically, the output from the previous CIF block, $\mathbf{F}_{c}^{i-1}$, is used as input for the current CIF block. $\mathbf{F}_{c}^{i-1}$ contains information extracted from the features of the previous backbone block's output, $\mathbf{F}_{vx}^{i-1}$. First, $\mathbf{F}_{c}^{i-1}$ passes through the mamba layer, which extracts the historical contextual information, resulting in $\mathbf{F}_{h}^{i}$. Simultaneously, the mamba layer updates its hidden states based on $\mathbf{F}_{c}^{i-1}$, thereby updating the contextual information. Next, $\mathbf{F}_{vx}^{i-1}$ is used as $Q$, while $\mathbf{F}_{h}^{i}$ serves as $K$ and $V$ in the in-attention layer to obtain $\mathbf{F}_{vx}^{{i-1}'}$, which incorporates the previous contextual information. Then, $\mathbf{F}_{vx}^{{i-1}'}$ is input into the backbone for further feature extraction, resulting in $\mathbf{F}_{vx}^{i}$. Subsequently, $\mathbf{F}_{vx}^{i}$ is used as $K$ and $V$, while $\mathbf{F}_{h}^{i}$ is used as $Q$ in the residual connected out-attention and FFN to extract the current frame's information into the CIF block, resulting in the current CIF block's output, $\mathbf{F}_{c}^{i}$. The processing of the CIF module can be summarized as follows:
\begin{equation}
\label{eq:cif}
\begin{aligned}
    & \mathbf{F}_{h}^{i}, \mathbf{H}_{t}^{i} = {\rm Mamba_{i}}(\mathbf{F}_{c}^{i-1}, \mathbf{H}_{t-1}^{i}), \\
    & \mathbf{F}_{vx}^{i} = {\rm Block_i}(\mathbf{F}_{vx}^{i-1} + {\rm Att_{in}^{i}}(\mathbf{F}_{vx}^{i-1},\mathbf{F}_{h}^{i},\mathbf{F}_{h}^{i})), \\
    & \mathbf{F}_{c}^{i'} = {\rm Att_{out}^i}(\mathbf{F}_h^{i},\mathbf{F}_{vx}^{i},\mathbf{F}_{vx}^{i}) + \mathbf{F}_h^{i}, \\
    & \mathbf{F}_{c}^{i} = \mathbf{F}_{c}^{i'} + {\rm FFN_{i}}(\mathbf{F}_{c}^{i'}),
\end{aligned}
\end{equation}
Where, ${\rm Mamba_{i}}$, ${\rm Att_{in}^{i}}$, ${\rm Att_{out}^i}$ and ${\rm FFN_{i}}$ represent the mamba layer, in-attention, out-attention, and FFN layer, respectively, in the $i_{th}$ CIF block, ${\rm Block_i}$ represents the $i_{th}$ backbone block, $\mathbf{H}_{t}^{i}$ and $\mathbf{H}_{t-1}^{i}$ represent the hidden states of the $i_{th}$ CIF block at time $t$ and ${t-1}$, respectively.

\begin{figure}[t]
\begin{center}
\includegraphics[width=1.0\linewidth]{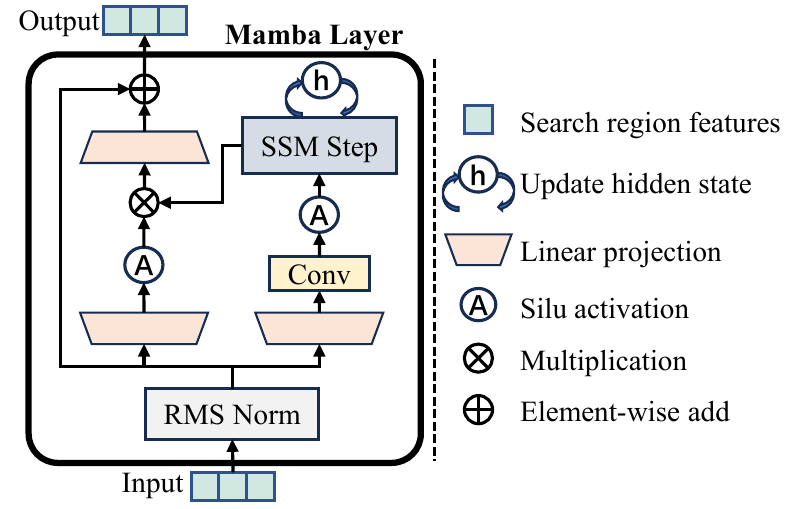}
\end{center}
   \caption{The detailed mamba layer in the CIF block.} 
\label{fig:mamba}
\end{figure}

\textit{Mamba Layer. }The mamba layer is a key component of the CIF block, responsible for storing contextual information. The detailed structure of the mamba layer is shown in Figure~\ref{fig:mamba}. First, the input $\mathbf{F}_i$ undergoes RMS Norm~\cite{rms} for normalization, resulting in $\mathbf{F}_{\text{in}}$. Next, $\mathbf{F}_{\text{in}}$ is processed through two branches: $x$ and $z$. In the $x$ branch, $\mathbf{F}_{\text{in}}$ is expanded in dimension via a linear projection and then passed through a convolutional layer to extract local features. After activation by the SiLU function~\cite{silu}, it is fed into the SSM to extract and update the stored contextual information. The output of the $x$ branch is denoted as $\mathbf{F}_x$. In the $z$ branch, $\mathbf{F}_{\text{in}}$ is dimensionally expanded and activated by the SiLU function. It is then weighted by $\mathbf{F}_x$ and reduced in dimension to produce the output $\mathbf{F}_i'$ of the $z$ branch. Finally, $\mathbf{F}_i'$ is combined with $\mathbf{F}_{\text{in}}$ through a skip connection to produce the output  $\mathbf{F}_o$ of the Mamba Layer. This output  $\mathbf{F}_o$ integrates the previously stored contextual information.

\textit{State Space Model. } The State Space Model (SSM) is a critical part of the mamba layer. Inspired by continuous systems, SSM maps the one-dimensional function $x(t)\in{\mathbb{R}}$ to  $y(t)\in{\mathbb{R}}$ through the system's hidden state function $h(t)\in {\mathbb{R}}^{N}$ ($N$ represents the state size). SSM can be represented by a set of linear ordinary differential equations, specifically the state equation and the output equation:

\begin{equation}
\label{eq:mamba}
\begin{aligned}
   h'(t) &= \textbf{A}h(t) + \textbf{B}x(t), \\
    y(t) &= \textbf{C}h(t) + \textbf{D}x(t),
\end{aligned}
\end{equation}
Where $\textbf{A} \in {\mathbb{R}}^{N\times N}$, $\textbf{B} \in {\mathbb{R}}^{N\times 1}$, $\textbf{C} \in {\mathbb{R}}^{1\times N}$, and $\textbf{D} \in  {\mathbb{R}}^1$ are the weighting parameters of the continuous-time SSM. To incorporate a continuous-time SSM in deep learning, it must be discretized. The zero-order hold method~\cite{mamba} is typically used to discretize it, converting continuous parameters into discrete parameters, as follows:

\begin{equation}
\begin{aligned}
    \mathbf{\bar{A}} &= \exp(\mathbf{\Delta A}), \\
    \mathbf{\bar{B}} &= (\mathbf{\Delta A})^{-1} (\exp(\mathbf{\Delta A}) - \mathbf{I}) \cdot \mathbf{\Delta B} \approx \mathbf{\Delta B},
\end{aligned}
\end{equation}
Where $\mathbf{\bar{A}}$ and $\mathbf{\bar{B}}$ are the discretized parameters, and $\mathbf{\Delta}$ is the time scale parameter. After discretization, the SSM used in the mamba layer  is represented by the following formula:
\begin{equation}
\label{eq:qmamba}
\begin{aligned}
   h_t &=  \mathbf{\bar{A}}h_{t-1} + \mathbf{\bar{B}}x_t, \\
    y_t &= \mathbf{C}h_t + \mathbf{D}x_t,
\end{aligned}
\end{equation}
Here, $h_{t-1}$ represents the previous hidden states that store crucial contextual information, $h_t$ is the updated hidden states based on the current input $x_t$, and $y_t$ is the output that integrates the contextual information. In this way,  SSM enables storing and transmitting contextual information.

\subsection{Head and Loss Function.}
We use the same classification and regression heads as OSTrack ~\cite{ostrack} for our prediction head.  The head consists of three sub-networks, each composed of convolutional layers.  These sub-networks produce the following outputs: the classification score $S\in {\mathbb{R}^{1\times {\frac{H_x}{16}}\times{\frac{W_x}{16}}}}$, bounding box size $B\in {\mathbb{R}^{2\times {\frac{H_x}{16}}\times{\frac{W_x}{16}}}}$, and offset size $O\in {\mathbb{R}^{2\times {\frac{H_x}{16}}\times{\frac{W_x}{16}}}}$. Using $S$, $B$, and $O$,  we predict the final tracking results. Our objective function includes classification loss, focal loss~\cite{focal_loss}, and  regression loss, which consists of L1 loss and GIoU loss~\cite{GIoU}.  In our video-level training approach, each frame contributes to an individual loss. The total loss can be summarized as follows:
\begin{equation}
	\label{equ-loss-dyhit}
	\begin{gathered}
		\mathcal{L} = \sum_{i=1}^{n}( \lambda_{c}\mathcal{L}_{cls}^{i} + \lambda_{l}\mathcal{L}_l^i+ \lambda_{g}\mathcal{L}_g^i),
	\end{gathered}
\end{equation}
Where, $\mathcal{L}_{cls}^{i}$, $\mathcal{L}_l^i$ and $\mathcal{L}_g^i$ represent the classification loss, L1 loss, and GIoU loss for the $i_{th}$ frame, respectively. $\mathcal{L}$ is the total loss. $\lambda_{c}$, $\lambda_{l}$ and $\lambda_{g}$ are hyperparameters, with default values $\lambda_{c} = 1$, $\lambda_{l} = 5$ and $\lambda_{g} = 2$.

\begin{table*}[ht]
  \centering
  \caption{State-of-the-art comparisons on four large-scale benchmarks. We add a symbol * over GOT-10k to indicate that the corresponding models are only trained with the GOT-10k training set. Otherwise, the models are trained with all the training data presented in Sec. Implementation Details.
  The top two results are highlighted with \textbf{bold} and \underline{underlined} fonts, respectively.}
  \label{tab-sota}
  \setlength{\tabcolsep}{1.5mm}{  
  \small
  \begin{tabular}{l| ccc c ccc c ccc c ccc}
    \toprule
    \multirow{2}*{Method} & \multicolumn{3}{c}{LaSOT}&& \multicolumn{3}{c}{LaSOT$_{ext}$} && \multicolumn{3}{c}{TrackingNet} && \multicolumn{3}{c}{GOT-10k*}\\
    \cline{2-4}
    \cline{6-8}
    \cline{10-12}
    \cline{14-16}
    & AUC&P$_{Norm}$&P && AUC&P$_{Norm}$&P && AUC&P$_{Norm}$&P && AO&SR$_{0.5}$&SR$_{0.75}$\\
    \midrule[0.5pt]
    MCITrack-B224	&\textbf{{75.3}}&\textbf{{85.6}}&\textbf{{83.3}}  & &\textbf{{54.6}} &\textbf{{65.7}} &\textbf{{62.1}}    & &\textbf{{86.3}}&\textbf{{90.9}}&\textbf{{86.1}}    & &\textbf{{77.9}} &\textbf{{88.2}} &\textbf{{76.8}}\\
    MCITrack-S224	&\underline{73.8}&\underline{84.2}&\underline{81.7} & &\underline{52.6} &63.6 &59.7     & &\underline{85.6}&\underline{90.2}	&\underline{85.2}  & &{76.9} &{87.0}&\underline{76.1}\\
    MCITrack-T224	&71.7&81.5&78.2  & &51.6&62.7&58.4      & &84.8&89.4&83.7    & &74.0&83.9&72.1\\
    \midrule[0.1pt]
    ODTrack-B384~\cite{odtrack} &{73.2}&{83.2}&{80.6}&     &{52.4}&\underline{63.9}&\underline{60.1}&    &{85.1}&{90.1}&{84.9}&   &\underline{77.0}&\underline{87.9}&{75.1}\\
    ARTrackV2-256~\cite{artrackv2} &71.6&80.2&77.2& &50.8&61.9&57.7& &84.9&89.3&84.5& &75.9&85.4&72.7 \\
    AQATrack-256~\cite{aqatrack} &71.4&81.9&78.6&   &51.2&62.2&58.9&    &83.8&88.6&83.1&      &73.8&83.2&72.1\\
    EVPTrack-224~\cite{evptrack} &70.4&80.9&77.2&     &48.7&59.5&55.1&   &83.5&88.3&- &          &73.3&83.6&70.7\\
    LoRAT-B224~\cite{lorat}&71.7&80.9&77.3&     &50.3&61.6&57.1&     &83.5&87.9&82.1&            &72.1&81.8&70.7\\
    ARTrack-256~\cite{artrack} &70.4&79.5&76.6&    &46.4&56.5&52.3&     &84.2&88.7&83.5&    &73.5&82.2&70.9\\
    SeqTrack-B256~\cite{seqtrack} &69.9&79.7&76.3&    &49.5&60.8&56.3&    &83.3&88.3&82.2&    &74.7&84.7&71.8\\
    VideoTrack-256~\cite{videotrack} &70.2&-&76.4 &&-&-&- &&83.8&88.7&83.1 &&72.9&81.9&69.8\\
    ROMTrack-384~\cite{romtrack} &71.4&81.4&78.2& &51.3&62.4&58.6& &84.1&89.0&83.7& &74.2&84.3&72.4\\
    CiteTracker-384~\cite{citetrack} &69.7&78.6&75.7& &-&-&-& &84.5&89.0&84.2& &74.7&84.3&73.0\\
    OSTrack-256~\cite{ostrack}	&69.1&78.7&75.2&      &47.4&57.3&53.3 &   &83.1 &87.8&82.0&      &71.0&80.4&68.2\\
    Mixformer-22k~\cite{mixformer}	&69.2	&78.7	&74.7 & &- &- &- & &83.1	&88.1	&81.6 & &70.7	&80.0	&67.8\\
    SwinTrack-384~\cite{swintrack}	&71.3	&-	&76.5 & &49.1 &- &55.6 & &84.0	&-	&82.8 &	&72.4	&-	&67.8\\
    \midrule[0.1pt] 
    \multicolumn{16}{c}{Comparison of Large-Scale Models}\\
        \midrule[0.1pt] 
    MCITrack-L384	&\textbf{{76.6}}	&\textbf{{86.1}}	&\textbf{{85.0}}&    &\underline{55.7} &\underline{66.5} &\underline{62.9} &     &\textbf{{87.9}}	&\textbf{{92.1}} & \textbf{{89.2}}&     & \textbf{{80.0}}&\underline{88.5}&\textbf{{80.2}} \\
    MCITrack-L224	&\underline{76.1}&\textbf{{86.1}}&\underline{84.1}  & &{54.8} &{65.6} &61.6    & &\underline{86.9} &\underline{91.3}&\underline{87.4}  & &{79.3} &\textbf{{89.3}} &{78.7}\\
    \midrule[0.1pt] 
    ODTrack-L384~\cite{odtrack}	&74.0&\underline{84.2}&{82.3}&     &53.9&65.4&{61.7} &   &{86.1} &{91.0}&{86.7}&  &78.2&87.2&77.3\\
    ARTrackV2-L384~\cite{artrackv2} &73.6&82.8&81.1&   &53.4&63.7&60.2&      &{86.1}&90.4&86.2&    &\underline{79.5}&{87.8}&\underline{79.6}\\
    LoRAT-L378~\cite{lorat}&{75.1}&{84.1}&82.0&     &\textbf{{56.6}}&\textbf{{69.0}}&\textbf{{65.1}}&     &85.6&89.7&85.4&            &77.5&86.2&{78.1}\\
    MixViT-L384~\cite{mixformer_journal} &72.4&82.2&80.1& &-&-&-& &85.4&90.2&85.7& &75.7&85.3&75.1\\
    ARTrack-L384~\cite{artrack} &73.1&82.2&80.3&    &52.8&62.9&59.7&     &85.6&89.6&86.0&    &78.5&87.4&77.8\\
    SeqTrack-L384~\cite{seqtrack} &72.5&81.5&79.3&    &50.7&61.6&57.5&    &85.5&89.8&85.8&    &74.8&81.9&72.2\\
    TATrack-L384~\cite{TATrack} &71.1&79.1&76.1& &-&-&-& &85.0&89.3&84.5&&-&-&-\\
    CTTrack-L320~\cite{CTTrack} &69.8&78.7&76.2& &-&-&-& &84.9&89.1&83.5& &72.8&81.3&71.5\\
  \bottomrule
\end{tabular}
}
\end{table*}

\section{Experiments}
\subsection{Implementation Details}
\label{subsec:Implementation Details}
\subsubsection{Model. }We develop five variants of the MCITrack model with different backbones and input resolutions, as detailed in Table~\ref{tab-model}. 
Additionally, we report the model parameters, FLOPs, and inference speed in Table~\ref{tab-model}. The speed is measured on an Intel Core i7-8700K CPU @3.70GHz with 47GB RAM and a single 2080 Ti GPU. All the models are implemented with Python 3.11 and PyTorch 2.1.2.

\subsubsection{Training. }Our training data includes LaSOT~\cite{lasot}, GOT-10k~\cite{GOT10K}, TrackingNet~\cite{trackingnet}, COCO~\cite{COCO}, and VastTrack~\cite{vasttrack}. We use a 5-frame video clip and two search regions as inputs. To simulate contextual information transmission, we first input the video clip and one search region, calculate the loss, and store the contextual information in hidden states. Then, we input the second search region and the previous frame's hidden states, calculate the loss, and sum the losses for backpropagation.
We use the AdamW~\cite{AdamW} optimizer with an initial learning rate of $4\times10^{-5}$ for the backbone and $4\times10^{-4}$ for the rest. The weight decay is set to $1\times10^{-4}$. The model is trained for a total of 300 epochs with $60k$ samples per epoch, and the learning rate decreases by a factor of 10 after 240 epochs. Training is performed on two 80GB Tesla A800 GPUs with a total batch size of 128.

\begin{table}[t]
\centering
\caption{Details of MCITrack model variants.}
\label{tab-model}
\setlength{\tabcolsep}{0.4mm}{
\small
\scalebox{0.96}{
\begin{tabular}{l| c c c c c}
\toprule
\multirow{2}{*}{Model}           & \multirow{2}{*}{~Backbone~} & ~Input~ & ~Params~ & ~FLOPs~ & ~Speed~ \\
~ &~ & Resolution & (M) & (G) & (\emph{fps}) \\
\midrule 
MCITrack-T224   &   Fast-iTPN-T      &   $224$$\times$$224$   &   32  &13   &  51  \\
MCITrack-S224  &   Fast-iTPN-S       &   $224$$\times$$224$   &   45  &19   &  40  \\
MCITrack-B224   &   Fast-iTPN-B      &   $224$$\times$$224$   &   88 &38    &  35  \\
MCITrack-L224   &   Fast-iTPN-L     &   $224$$\times$$224$   &   287 &123    &  16  \\
MCITrack-L384  &   Fast-iTPN-L     &   $384$$\times$$384$   &   287 &370    &  5  \\
\bottomrule
\end{tabular}}
}
\end{table}

\subsubsection{Inference. }During inference, we input a video clip of length 5, consistent with our training process. We maintain a memory bank to store reliable frames. When the update interval $T$ is reached, we use the frames stored in the memory bank to update the video clip. We observe that updating the hidden states when tracking results are inaccurate can introduce erroneous contextual information, which may degrade the model's performance in subsequent frames. To mitigate this issue, we set a threshold $a$ for updating the hidden states: the hidden states are only updated if  the classification score exceeds the threshold. This approach helps prevent the model from being misled by incorrect information.
\subsection{State-of-the-Art Comparisons}
We compare our MCITrack with SOTA trackers across eight tracking benchmarks, as detailed in Tables~\ref{tab-sota} and~\ref{tab-sota-small}.


\subsubsection{LaSOT. }LaSOT~\cite{lasot} is a large-scale, long-term dataset with 1120 training videos and  208 test videos. As shown in Table~\ref{tab-sota}, MCITrack models achieve state-of-the-art results. Specifically, MCITrack-B224 and MCITrack-S224 attain  the top two AUC scores of 75.3\% and 73.8\%, respectively, outperforming the third-place ODTrack-B384 by 2.1\% and 0.6\%. Furthermore, among the large-scale models, MCITrack-L384 and MCITrack-L224 secure  the first and second positions with AUC scores of 76.6\% and 76.1\%. 

\begin{table}[t]
    \caption{Comparison with state-of-the-art methods on additional benchmarks in AUC score.}
\label{tab-sota-small}
  \centering
 \setlength{\tabcolsep}{1.5mm}{
    \small
    \begin{tabular}{l|ccc}
    \toprule
    Method &TNL2K&NFS&UAV123\\
    \midrule[0.5pt]
    MCITrack-L384 &\textbf{{65.3}}	 &\underline{70.6}	&\underline{71.5}	 \\
    MCITrack-L224 &\underline{64.3}	&\textbf{{71.1}}	&70.8\\
    MCITrack-B224 &{62.9}&\underline{70.6}	&70.5 \\
    MCITrack-S224 &61.9&\underline{70.6}	&69.3 \\
    MCITrack-T224 &59.4&{70.0}&69.9 \\
    \midrule[0.1pt]
    ODTrack-L384~\cite{odtrack} &61.7&-&- \\
    ARTrackV2-256~\cite{artrackv2} &59.2&67.6&69.9 \\
    LoRAT-L378~\cite{lorat} &62.3&66.7&\textbf{{72.5}}	\\
    ARTrack-L384~\cite{artrack} &60.3 &67.9 &{71.2}\\
    SeqTrack-L384~\cite{seqtrack} &57.8 &66.2 &68.5 \\
    OSTrack~\cite{ostrack}&55.9&66.5&70.7\\
    STARK~\cite{Stark}&-&66.2&68.2 \\
    TransT~\cite{transt}&50.7&65.7&69.1 \\
    DiMP~\cite{DiMP}&44.7&61.8&64.3 \\
    Ocean~\cite{ocean}&38.4&49.4&57.4 \\
    \bottomrule
    \end{tabular}
    }
\end{table}

\subsubsection{LaSOT$_{ext}$.}LaSOT$_{ext}$~\cite{lasot_journal} is an extension of LaSOT, containing 150 videos. As reported in Table~\ref{tab-sota}, our models achieve the top two results. MCITrack-B224 achieves an AUC score of 54.6\%, which is 2.2\% higher than the previous best, ODTrack-B384, representing a significant improvement. With a large-scale backbone, MCITrack-L384 achieves a competitive AUC score of 55.7\%.
\subsubsection{TrackingNet. }TrackingNet~\cite{trackingnet} is a comprehensive  dataset that covers diverse object categories and scenes. As presented in Table~\ref{tab-sota}, MCITrack-B224 and MCITrack-S224 achieve the top two positions with AUC scores of 86.3\% and 85.6\%, respectively. Additionally, MCITrack-L384 and MCITrack-L224 also secure the top two results among large-scale trackers, with AUC scores of 1.8\% and 0.8\% higher than the third-place tracker, demonstrating a significant performance advantage.
\subsubsection{GOT-10k. }GOT-10k~\cite{GOT10K} test set contains 180 videos covering various common tracking challenges. In line with official guidelines, we only use the GOT-10k training set for model training. As shown in Table~\ref{tab-sota}, MCITrack-B224 achieves the highest AO score of 77.9\%, surpassing the second-place ODTrack-B384 by 0.9\%. Additionally, among the large-scale models, MCITrack-L384 achieves the top AO score of 80.0\%.

\begin{figure}[t]
\begin{center}
\includegraphics[width=1\linewidth]{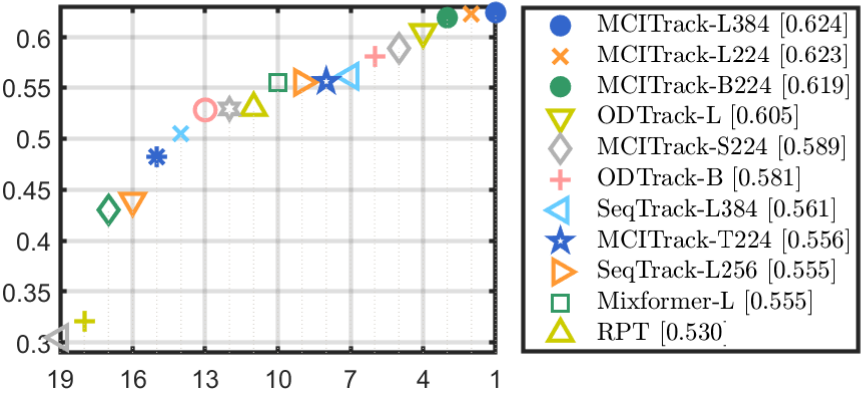}
\end{center}
   \caption{EAO rank plots on VOT2020.} 
\label{fig:vot}
\end{figure}
\subsubsection{TNL2K, NFS and UAV123. }We also evaluate our MCITrack models on three additional datasets:  TNL2K~\cite{TNL2K}, NFS~\cite{NFS}, and UAV123~\cite{UAV}. TNL2K is a recently released large-scale dataset with 700 video sequences, while NFS and UAV123 are two smaller benchmarks, containing 100 and 123 videos, respectively. As shown in Table~\ref{tab-sota-small}, MCITrack achieves new state-of-the-art performance on TNL2K and NFS, with AUC scores of 65.3\% and 71.1\%, respectively. On UAV123, MCITrack-L384 also delivers a competitive result, achieving an AUC score of 71.5\%, which is better than most previous trackers.
\subsubsection{VOT2020. }VOT2020~\cite{vot2020} contains 60 challenging videos. We equip MCITrack with Alpha-Refine~\cite{Alpha-Refine} to predict segmentation masks. As shown in Figure~\ref{fig:vot}, MCITrack-L384, MCITrack-L224, and MCITrack-B224 secure the top three positions with EAO scores of 62.4\%, 62.3\%, and 61.9\%, respectively.

\subsection{Ablation Study}
In the ablation study, we use MCITrack-B224 as our baseline model to investigate the impact of the CIF module, video clip length, and different methods of contextual information transmission on the model.
\subsubsection{CIF Module. }The CIF module is a crucial component of MCITrack, responsible for storing and transmitting contextual information. We conduct ablation studies on the number of CIF blocks in the CIF module, the structure of the CIF block, and the size of the hidden state. The results are shown in Table~\ref{tab-ablation}. Our baseline model uses 4 CIF blocks with a hidden state size of 16. As observed (\#2 and \#3), decreasing or increasing the number of CIF blocks to 2 or 6 results in a reduction in AUC on LaSOT by 1.4\% and 0.5\%, respectively. This demonstrates the effectiveness of our method for deep integration with the backbone, while too many interactions increase model complexity and can degrade performance. We also explore different structures of the CIF block, \#4, \#5, and \#6 show the results. By replacing the attention modules with addition, we find that removing in-attention, out-attention, or both reduces the model performance by 1.0\%, 1.1\%, and 1.8\%, respectively. These results underscore the effectiveness of our CIF block structure. Table~\ref{tab-ablation} (\#7, \#8, and \#9) also demonstrates the impact of different hidden state sizes. Both too small and too large hidden state sizes lead to decreased model performance, indicating the importance of optimal hidden state size for maintaining performance.

\begin{table}[t]
\definecolor{purple(x11)}{rgb}{0.63, 0.36, 0.94}
\definecolor{yellow(munsell)}{rgb}{1.0,0.988, 0.957}
\definecolor{green(colorwheel)(x11green)}{rgb}{0.0, 1.0, 0.0}
\definecolor{pink}{rgb}{1.0, 0.85, 0.85}
\centering
\caption{Ablation Study of the CIF module  on LaSOT. 
\#1 denotes the baseline setting. \#2 and \#3 denote the  number of CIF blocks. \#4, \#5, and \#6 denote the CIF block structure. \#7, \#8, and \#9 denote the hidden state size.  $\Delta$ denotes the performance change (AUC) compared with the baseline.}
\label{tab-ablation}
\small
\setlength{\tabcolsep}{1mm}{
\begin{tabular}{l|c|ccc|c}
\toprule
\# & Method &AUC &P$_{Norm}$&P&$\Delta$\\
\midrule

1 & Baseline &75.3 &85.6 &83.3 &-\\

2 &2 CIF blocks &73.9 &84.5 &81.8 &\textbf{-1.4}\\  

3 &6 CIF blocks &74.8 &84.8 &82.8 &\textbf{-0.5}\\  

4 &Without in cross attention &74.3 &84.3 &81.8 &\textbf{-1.0}\\

5 &Without out cross attention &74.2 &84.1 &82.0 &\textbf{-1.1}\\

6 &Without in \& out cross attention &73.5 &83.3 &80.7 &\textbf{-1.8}\\

7 &4 hidden states &74.0 &83.9 &81.7  &\textbf{-1.3}\\

8 &8 hidden states &74.6 &84.8 &82.7  &\textbf{-0.7}\\

9 &32 hidden states &74.3 &84.2 &81.9  &\textbf{-1.0}\\
\bottomrule
\end{tabular}
}
\end{table}

\begin{table}[t]
    \caption{Ablation Study of the video clip length on LaSOT.    $\Delta$ denotes the performance change (AUC) compared with the baseline.}
\label{tab-ablation-seql}
  \centering
  \setlength{\tabcolsep}{3mm}{
    \small
    \begin{tabular}{c|ccc|c}
    \toprule
    Video Clip Length &AUC&P$_{Norm}$&P &$\Delta$\\
    \midrule[0.5pt]
    Baseline &75.3 &85. 6&83.3 & - \\
    2 frames &72.7&82.4&80.0 &\textbf{-2.6}\\
    3 frames &73.8&83.5&81.3 &\textbf{-1.5} \\
    4 frames &74.4&84.2&81.9 &\textbf{-0.9} \\
    6 frames &73.8&83.8&81.3 &\textbf{-1.5} \\
    \bottomrule
    \end{tabular}
    }
\end{table}

\begin{table}[t]
    \caption{Ablation Study of different contextual information propagation methods on LaSOT. $\Delta$ denotes the performance change (AUC) compared with the baseline.}
\label{tab-ablation-cp}
  \centering
  \setlength{\tabcolsep}{3mm}{
    \small
    \begin{tabular}{c|ccc|c}
    \toprule
    Method &AUC&P$_{Norm}$&P &$\Delta$\\
    \midrule[0.5pt]
    Baseline &75.3 &85.6&83.3 & - \\
    Wo CI  &73.0&83.0&80.6 &\textbf{-2.3}\\
    Extra Token &73.4&83.0&80.7 &\textbf{-1.9} \\
    Previous Features &73.4&83.2&80.6 &\textbf{-1.9} \\
    LSTM&74.3&84.2&81.8 &\textbf{-1.0} \\
    \bottomrule
    \end{tabular}
    }
  
\end{table}

\subsubsection{Video Clip Length. }Table~\ref{tab-ablation-seql} shows the impact of video clip length on MCITrack. Our baseline uses a 5-frame clip. The results reveal that as the video clip length increases, incorporating more appearance information, the model performance also improves, with an increase of 2.6\% in AUC on LaSOT when extending from 2 frames to 5 frames. However, further increasing the video clip length does not consistently lead to performance gains and may even cause performance degradation. For instance, using 6 frames results in a 1.5\% decrease in AUC on LaSOT. This suggests that excessively long video clips can impose a higher learning burden on the model, potentially leading to reduced performance.
\subsubsection{Contextual Information Propagation Methods. }Table~\ref{tab-ablation-cp} compares different methods for transmitting contextual information. The baseline is MCITrack, which uses Mamba's hidden states to transmit contextual information. "Wo CI" indicates the absence of contextual information in tracking. "Extra Token" refers to use additional tokens for transmitting contextual information. "LSTM" indicates use LSTM's hidden states for this purpose, while "Previous Features" involves using features from the previous frame directly. Compared to the baseline, "Wo CI" results in a 2.3\% decrease in AUC on LaSOT, highlighting the importance of contextual information for the tracker. "Extra Token" and "Previous Features" result in a 1.9\% decrease in AUC compared to the baseline. In contrast, "LSTM" achieves a relatively higher AUC of 74.3\%, which is 0.9\% better than "Extra Token" and "Previous Features". This suggests that our method of using hidden states to transmit contextual information is effective. Compared with previous methods, our method can convey richer and more crucial contextual information.
\subsubsection{Visualization. }
\begin{figure}[t]
\begin{center}
\includegraphics[width=1\linewidth]{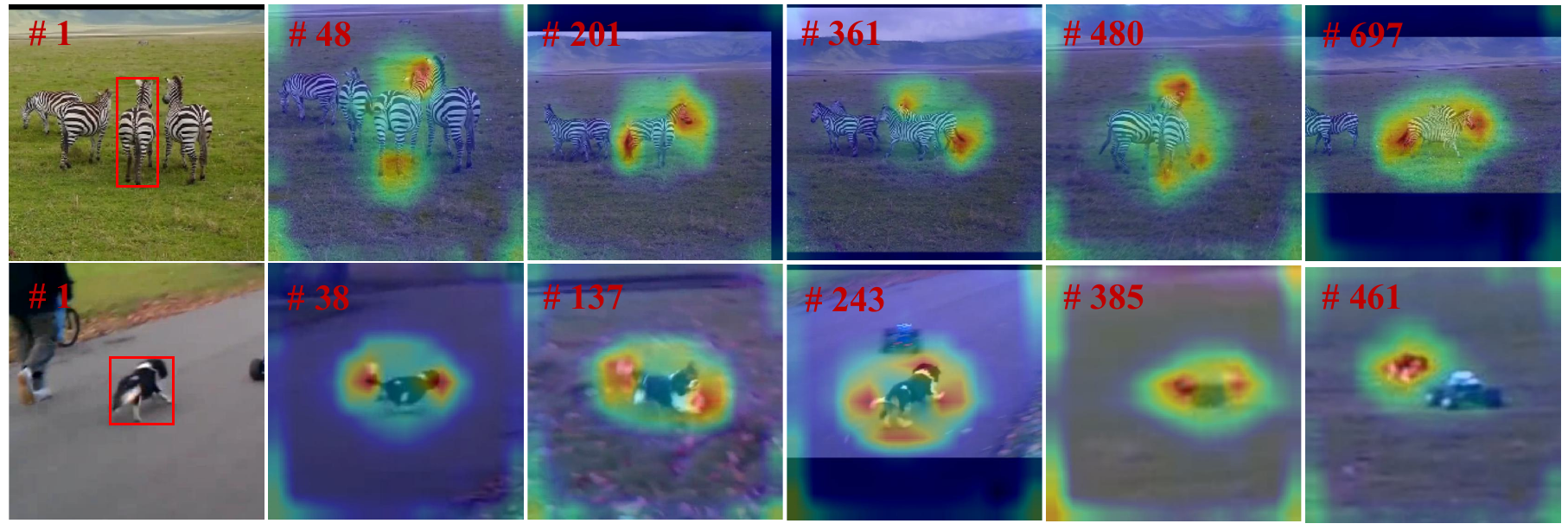}
\end{center}
   \caption{Visualization of the output features of CIF module.} 
\label{fig:vis}
\end{figure}

Figure~\ref{fig:vis} presents the visualization results of the output features from the CIF module. It shows that, after passing through the CIF module, the model emphasizes critical edge information of the target, such as the tail, legs, and head of the zebra and the dog. This enhancement allows for more accurate target localization. 

\section{Conclusion}
In this work, we introduce a new family of video-level tracking models named MCITrack, which develops a new method for transmitting contextual information. Its core component is the Contextual Information Fusion(CIF) module, which utilizes hidden states from Mamba to effectively transmit richer and more crucial contextual information. Experiments show that  MCITrack is effective, achieving state-of-the-art performance across multiple datasets. We hope this work inspires further research in video-level tracking.

\textit{Limitation. }Despite its significant results, MCITrack faces challenges with slow model training due to its video-level modeling approach. Additionally,  handling the video clip introduces more computational overhead compared to the single-frame image. To address these issues, it is essential to accelerate model training and minimize the computational burden of the video clip. A promising solution is to efficiently integrate multi-frame image information to reduce the video clip size and develop a lightweight model.

\section{Acknowledgements}
The paper is supported in part by National Natural Science Foundation of China (nos. U23A20384 and 62476044), in part by Key Research and Development Program of Liaoning Province (no. 2023JH26/10200015), and in part by Open Research Fund of the National Key Laboratory of Space Intelligent Control Technology (No. HTKJ2024KL502016).

\bibliography{aaai25}

\begin{thebibliography}{67}
\providecommand{\natexlab}[1]{#1}

\bibitem[{Bai et~al.(2024)Bai, Zhao, Gong, and Wei}]{artrackv2}
Bai, Y.; Zhao, Z.; Gong, Y.; and Wei, X. 2024.
\newblock ARTrackV2: Prompting Autoregressive Tracker Where to Look and How to Describe.
\newblock In \emph{CVPR}, 19048--19057.

\bibitem[{Bertinetto et~al.(2016)Bertinetto, Valmadre, Henriques, Vedaldi, and Torr}]{SiameseFC}
Bertinetto, L.; Valmadre, J.; Henriques, J.~F.; Vedaldi, A.; and Torr, P. H.~S. 2016.
\newblock Fully-Convolutional Siamese Networks for Object Tracking.
\newblock In \emph{ECCV}, 850--865.

\bibitem[{Bhat et~al.(2019)Bhat, Danelljan, Gool, and Timofte}]{DiMP}
Bhat, G.; Danelljan, M.; Gool, L.~V.; and Timofte, R. 2019.
\newblock Learning Discriminative Model Prediction for Tracking.
\newblock In \emph{ICCV}, 6182--6191.

\bibitem[{Cai et~al.(2023)Cai, Liu, Tang, and Wu}]{romtrack}
Cai, Y.; Liu, J.; Tang, J.; and Wu, G. 2023.
\newblock Robust Object Modeling for Visual Tracking.
\newblock In \emph{ICCV}, 9589--9600.

\bibitem[{Cao et~al.(2022)Cao, Huang, Pan, Zhang, Liu, and Fu}]{tcttrack}
Cao, Z.; Huang, Z.; Pan, L.; Zhang, S.; Liu, Z.; and Fu, C. 2022.
\newblock TCTrack: Temporal Contexts for Aerial Tracking.
\newblock In \emph{CVPR}, 14798--14808.

\bibitem[{Chang et~al.(2024)Chang, Yongsheng, Xin, Huchuan, and Dong}]{stid}
Chang, L.; Yongsheng, Y.; Xin, C.; Huchuan, L.; and Dong, W. 2024.
\newblock Spatial-temporal initialization dilemma: towards realistic visual tracking.
\newblock \emph{Visual Intelligence 2, Article no. 35.}

\bibitem[{Chen et~al.(2022)Chen, Li, Bai, Qiao, Shen, Li, Gan, Wu, and Ouyang}]{simtrack}
Chen, B.; Li, P.; Bai, L.; Qiao, L.; Shen, Q.; Li, B.; Gan, W.; Wu, W.; and Ouyang, W. 2022.
\newblock Backbone is All Your Need: A Simplified Architecture for Visual Object Tracking.
\newblock In \emph{ECCV}, 375--392.

\bibitem[{Chen et~al.(2023{\natexlab{a}})Chen, Peng, Wang, Lu, and Hu}]{seqtrack}
Chen, X.; Peng, H.; Wang, D.; Lu, H.; and Hu, H. 2023{\natexlab{a}}.
\newblock SeqTrack: Sequence to Sequence Learning for Visual Object Tracking.
\newblock In \emph{CVPR}, 14572--14581.

\bibitem[{Chen et~al.(2023{\natexlab{b}})Chen, Yan, Zhu, Lu, Ruan, and Wang}]{transt_journal}
Chen, X.; Yan, B.; Zhu, J.; Lu, H.; Ruan, X.; and Wang, D. 2023{\natexlab{b}}.
\newblock High-Performance Transformer Tracking.
\newblock \emph{IEEE TPAMI}, 8507--8523.

\bibitem[{Chen et~al.(2021)Chen, Yan, Zhu, Wang, Yang, and Lu}]{transt}
Chen, X.; Yan, B.; Zhu, J.; Wang, D.; Yang, X.; and Lu, H. 2021.
\newblock Transformer Tracking.
\newblock In \emph{CVPR}, 8126--8135.

\bibitem[{Chen et~al.(2020)Chen, Zhong, Li, Zhang, and Ji}]{SiamBAN}
Chen, Z.; Zhong, B.; Li, G.; Zhang, S.; and Ji, R. 2020.
\newblock Siamese Box Adaptive Network for Visual Tracking.
\newblock In \emph{CVPR}, 6668--6677.

\bibitem[{Cui et~al.(2022)Cui, Jiang, Wang, and Wu}]{mixformer}
Cui, Y.; Jiang, C.; Wang, L.; and Wu, G. 2022.
\newblock MixFormer: End-to-End Tracking with Iterative Mixed Attention.
\newblock In \emph{CVPR}, 13608--13618.

\bibitem[{Cui et~al.(2024)Cui, Jiang, Wang, and Wu}]{mixformer_journal}
Cui, Y.; Jiang, C.; Wang, L.; and Wu, G. 2024.
\newblock MixFormer: End-to-End Tracking with Iterative Mixed Attention.
\newblock \emph{IEEE TPAMI}, 0--18.

\bibitem[{Danelljan et~al.(2019)Danelljan, Bhat, Khan, and Felsberg}]{ATOM}
Danelljan, M.; Bhat, G.; Khan, F.~S.; and Felsberg, M. 2019.
\newblock {ATOM: Accurate} Tracking by Overlap Maximization.
\newblock In \emph{CVPR}, 4660--4669.

\bibitem[{Dosovitskiy et~al.(2020)Dosovitskiy, Beyer, Kolesnikov, Weissenborn, Zhai, Unterthiner, Dehghani, Minderer, Heigold, Gelly et~al.}]{ViT}
Dosovitskiy, A.; Beyer, L.; Kolesnikov, A.; Weissenborn, D.; Zhai, X.; Unterthiner, T.; Dehghani, M.; Minderer, M.; Heigold, G.; Gelly, S.; et~al. 2020.
\newblock An Image is Worth 16x16 Words: Transformers for Image Recognition at Scale.
\newblock In \emph{ICLR}, 1--9.

\bibitem[{Elfwing, Uchibe, and Doya(2018)}]{silu}
Elfwing, S.; Uchibe, E.; and Doya, K. 2018.
\newblock Sigmoid-Weighted Linear Units for Neural Network Function Approximation in Reinforcement Learning.
\newblock \emph{Neural Networks}, 3--11.

\bibitem[{Fan et~al.(2021)Fan, Bai, Lin, Yang, Chu, Deng, Yu, Huang, Liu, Xu et~al.}]{lasot_journal}
Fan, H.; Bai, H.; Lin, L.; Yang, F.; Chu, P.; Deng, G.; Yu, S.; Huang, M.; Liu, J.; Xu, Y.; et~al. 2021.
\newblock LaSOT: A High-Quality Large-Scale Single Object Tracking Benchmark.
\newblock \emph{IJCV}, 439--461.

\bibitem[{Fan et~al.(2019)Fan, Lin, Yang, Chu, Deng, Yu, Bai, Xu, Liao, and Ling}]{lasot}
Fan, H.; Lin, L.; Yang, F.; Chu, P.; Deng, G.; Yu, S.; Bai, H.; Xu, Y.; Liao, C.; and Ling, H. 2019.
\newblock {LaSOT}: A High-Quality Benchmark for Large-Scale Single Object Tracking.
\newblock In \emph{CVPR}, 5374--5383.

\bibitem[{Fu et~al.(2021)Fu, Liu, Fu, and Wang}]{stmtrack}
Fu, Z.; Liu, Q.; Fu, Z.; and Wang, Y. 2021.
\newblock STMTrack: Template-free Visual Tracking with Space-time Memory Networks.
\newblock In \emph{CVPR}, 13774--13783.

\bibitem[{Gao et~al.(2022)Gao, Zhou, Ma, Wang, and Yuan}]{AiATrack}
Gao, S.; Zhou, C.; Ma, C.; Wang, X.; and Yuan, J. 2022.
\newblock {AiATrack}: Attention in Attention for Transformer Visual Tracking.
\newblock In \emph{ECCV}, 146--164.

\bibitem[{Graves(2012)}]{lstm}
Graves, A. 2012.
\newblock \emph{Supervised Sequence Labelling with Recurrent Neural Networks}.
\newblock Studies in Computational Intelligence. Springer.

\bibitem[{Gu and Dao(2023)}]{mamba}
Gu, A.; and Dao, T. 2023.
\newblock Mamba: Linear-Time Sequence Modeling with Selective State Spaces.
\newblock \emph{arXiv preprint arXiv:2312.00752}.

\bibitem[{Gu, Goel, and R{\'{e}}(2022)}]{ssm1}
Gu, A.; Goel, K.; and R{\'{e}}, C. 2022.
\newblock Efficiently Modeling Long Sequences with Structured State Spaces.
\newblock In \emph{ICLR}, 1--9.

\bibitem[{Gu et~al.(2021)Gu, Johnson, Goel, Saab, Dao, Rudra, and R{\'e}}]{ssm}
Gu, A.; Johnson, I.; Goel, K.; Saab, K.; Dao, T.; Rudra, A.; and R{\'e}, C. 2021.
\newblock Combining Recurrent, Convolutional, and Continuous-Time Models with Linear State Space Layers.
\newblock In \emph{NeurIPS}, 572--585.

\bibitem[{Guo et~al.(2022)Guo, Zhang, Fan, Jing, Lyu, Li, and Hu}]{transinmo}
Guo, M.; Zhang, Z.; Fan, H.; Jing, L.; Lyu, Y.; Li, B.; and Hu, W. 2022.
\newblock Learning Target-aware Representation for Visual Tracking via Informative Interactions.
\newblock In \emph{IJCAI}, 927--934.

\bibitem[{He et~al.(2023)He, Zhang, Xie, Li, and Wang}]{TATrack}
He, K.; Zhang, C.; Xie, S.; Li, Z.; and Wang, Z. 2023.
\newblock Target-Aware Tracking with Long-Term Context Attention.
\newblock In \emph{AAAI}, 773--780.

\bibitem[{Huang, Zhao, and Huang(2019)}]{GOT10K}
Huang, L.; Zhao, X.; and Huang, K. 2019.
\newblock GOT-10k: A Large High-Diversity Benchmark for Generic Object Tracking in The Wild.
\newblock \emph{IEEE TPAMI}, 1562--1577.

\bibitem[{Kiani~Galoogahi et~al.(2017)Kiani~Galoogahi, Fagg, Huang, Ramanan, and Lucey}]{NFS}
Kiani~Galoogahi, H.; Fagg, A.; Huang, C.; Ramanan, D.; and Lucey, S. 2017.
\newblock Need for Speed: A Benchmark for Higher Frame Rate Object Tracking.
\newblock In \emph{ICCV}, 1125--1134.

\bibitem[{Kristan et~al.(2020)Kristan, Leonardis, Matas, Felsberg, Pflugfelder, K{\"a}m{\"a}r{\"a}inen, Danelljan, Zajc, Luke{\v{z}}i{\v{c}}, Drbohlav et~al.}]{vot2020}
Kristan, M.; Leonardis, A.; Matas, J.; Felsberg, M.; Pflugfelder, R.; K{\"a}m{\"a}r{\"a}inen, J.-K.; Danelljan, M.; Zajc, L.~{\v{C}}.; Luke{\v{z}}i{\v{c}}, A.; Drbohlav, O.; et~al. 2020.
\newblock The Eighth Visual Object Tracking {VOT}2020 Challenge Results.
\newblock In \emph{ECCV}, 547--601.

\bibitem[{Li et~al.(2019)Li, Wu, Wang, Zhang, Xing, and Yan}]{SiamRPNplusplus}
Li, B.; Wu, W.; Wang, Q.; Zhang, F.; Xing, J.; and Yan, J. 2019.
\newblock {SiamRPN++}: {Evolution} of Siamese Visual Tracking with Very Deep Networks.
\newblock In \emph{CVPR}, 4282--4291.

\bibitem[{Li et~al.(2018)Li, Yan, Wu, Zhu, and Hu}]{SiameseRPN}
Li, B.; Yan, J.; Wu, W.; Zhu, Z.; and Hu, X. 2018.
\newblock High Performance Visual Tracking With Siamese Region Proposal Network.
\newblock In \emph{CVPR}, 8971--8980.

\bibitem[{Li et~al.(2023)Li, Huang, He, Wang, Lu, and Yang}]{citetrack}
Li, X.; Huang, Y.; He, Z.; Wang, Y.; Lu, H.; and Yang, M.-H. 2023.
\newblock Citetracker: Correlating Image and Text for Visual Tracking.
\newblock In \emph{ICCV}, 9974--9983.

\bibitem[{Lin et~al.(2022)Lin, Fan, Xu, and Ling}]{swintrack}
Lin, L.; Fan, H.; Xu, Y.; and Ling, H. 2022.
\newblock Swintrack: A Simple and Strong Baseline for Transformer Tracking.
\newblock In \emph{NeurIPS}, 16743--16754.

\bibitem[{Lin et~al.(2024)Lin, Fan, Zhang, Wang, Xu, and Ling}]{lorat}
Lin, L.; Fan, H.; Zhang, Z.; Wang, Y.; Xu, Y.; and Ling, H. 2024.
\newblock Tracking Meets LoRA: Faster Training, Larger Model, Stronger Performance.
\newblock In \emph{ECCV}, 1--15.

\bibitem[{Lin et~al.(2017)Lin, Goyal, Girshick, He, and Doll{\'a}r}]{focal_loss}
Lin, T.-Y.; Goyal, P.; Girshick, R.; He, K.; and Doll{\'a}r, P. 2017.
\newblock Focal Loss for Dense Object Detection.
\newblock In \emph{ICCV}, 2980--2988.

\bibitem[{Lin et~al.(2014)Lin, Maire, Belongie, Bourdev, Girshick, Hays, Perona, Ramanan, Doll{\'a}r, and Zitnick}]{COCO}
Lin, T.-Y.; Maire, M.; Belongie, S.~J.; Bourdev, L.~D.; Girshick, R.~B.; Hays, J.; Perona, P.; Ramanan, D.; Doll{\'a}r, P.; and Zitnick, C.~L. 2014.
\newblock {Microsoft COCO}: Common Objects in Context.
\newblock In \emph{ECCV}, 740--755.

\bibitem[{Liu et~al.(2024)Liu, Tian, Zhao, Yu, Xie, Wang, Ye, and Liu}]{vmamba}
Liu, Y.; Tian, Y.; Zhao, Y.; Yu, H.; Xie, L.; Wang, Y.; Ye, Q.; and Liu, Y. 2024.
\newblock VMamba: Visual State Space Model.
\newblock \emph{arXiv preprint arXiv:2401.10166}.

\bibitem[{Loshchilov and Hutter(2018)}]{AdamW}
Loshchilov, I.; and Hutter, F. 2018.
\newblock Decoupled Weight Decay Regularization.
\newblock In \emph{ICLR}, 1--9.

\bibitem[{Mayer et~al.(2022)Mayer, Danelljan, Bhat, Paul, Paudel, Yu, and Van~Gool}]{ToMP}
Mayer, C.; Danelljan, M.; Bhat, G.; Paul, M.; Paudel, D.~P.; Yu, F.; and Van~Gool, L. 2022.
\newblock Transforming Model Prediction for Tracking.
\newblock In \emph{CVPR}, 8731--8740.

\bibitem[{Mayer et~al.(2021)Mayer, Danelljan, Paudel, and Van~Gool}]{keeptrack}
Mayer, C.; Danelljan, M.; Paudel, D.~P.; and Van~Gool, L. 2021.
\newblock Learning Target Candidate Association to Keep Track of What Not to Track.
\newblock In \emph{ICCV}, 13444--13454.

\bibitem[{Mueller, Smith, and Ghanem(2016)}]{UAV}
Mueller, M.; Smith, N.; and Ghanem, B. 2016.
\newblock A Benchmark and Simulator for {UAV} Tracking.
\newblock In \emph{ECCV}, 445--461.

\bibitem[{Muller et~al.(2018)Muller, Bibi, Giancola, Alsubaihi, and Ghanem}]{trackingnet}
Muller, M.; Bibi, A.; Giancola, S.; Alsubaihi, S.; and Ghanem, B. 2018.
\newblock Tracking{N}et: A Large-Scale Dataset and Benchmark for Object Tracking in The Wild.
\newblock In \emph{ECCV}, 300--317.

\bibitem[{Peng et~al.(2024)Peng, Gao, Liu, Li, Dong, Zhang, Fan, and Zhang}]{vasttrack}
Peng, L.; Gao, J.; Liu, X.; Li, W.; Dong, S.; Zhang, Z.; Fan, H.; and Zhang, L. 2024.
\newblock VastTrack: Vast Category Visual Object Tracking.
\newblock \emph{arXiv preprint arXiv:2403.03493}.

\bibitem[{Rezatofighi et~al.(2019)Rezatofighi, Tsoi, Gwak, Sadeghian, Reid, and Savarese}]{GIoU}
Rezatofighi, H.; Tsoi, N.; Gwak, J.; Sadeghian, A.; Reid, I.~D.; and Savarese, S. 2019.
\newblock Generalized Intersection Over Union: {A} Metric and a Loss for Bounding Box Regression.
\newblock In \emph{CVPR}, 658--666.

\bibitem[{Ronneberger, Fischer, and Brox(2015)}]{unet}
Ronneberger, O.; Fischer, P.; and Brox, T. 2015.
\newblock U-Net: Convolutional Networks for Biomedical Image Segmentation.
\newblock In \emph{MICCAI}, 234--241.

\bibitem[{Shi et~al.(2024)Shi, Zhong, Liang, Li, Zhang, and Li}]{evptrack}
Shi, L.; Zhong, B.; Liang, Q.; Li, N.; Zhang, S.; and Li, X. 2024.
\newblock Explicit Visual Prompts for Visual Object Tracking.
\newblock In \emph{AAAI}, 4838--4846.

\bibitem[{Song et~al.(2023)Song, Luo, Yu, Chen, and Yang}]{CTTrack}
Song, Z.; Luo, R.; Yu, J.; Chen, Y.-P.~P.; and Yang, W. 2023.
\newblock Compact Transformer Tracker with Correlative Masked Modeling.
\newblock In \emph{AAAI}, 2321--2329.

\bibitem[{Tao, Gavves, and Smeulders(2016)}]{SINT}
Tao, R.; Gavves, E.; and Smeulders, A. W.~M. 2016.
\newblock Siamese Instance Search for Tracking.
\newblock In \emph{CVPR}, 1420--1429.

\bibitem[{Tian et~al.(2024)Tian, Xie, Qiu, Jiao, Wang, Tian, and Ye}]{fastitpn}
Tian, Y.; Xie, L.; Qiu, J.; Jiao, J.; Wang, Y.; Tian, Q.; and Ye, Q. 2024.
\newblock Fast-iTPN: Integrally Pre-Trained Transformer Pyramid Network with Token Migration.
\newblock \emph{IEEE TPAMI}, 1--15.

\bibitem[{Voigtlaender et~al.(2020)Voigtlaender, Luiten, Torr, and Leibe}]{SiamRCNN}
Voigtlaender, P.; Luiten, J.; Torr, P. H.~S.; and Leibe, B. 2020.
\newblock Siam {R-CNN:} {V}isual Tracking by Re-Detection.
\newblock In \emph{CVPR}, 6578--6588.

\bibitem[{Wang et~al.(2021{\natexlab{a}})Wang, Zhou, Wang, and Li}]{TMT}
Wang, N.; Zhou, W.; Wang, J.; and Li, H. 2021{\natexlab{a}}.
\newblock Transformer Meets Tracker: Exploiting Temporal Context for Robust Visual Tracking.
\newblock In \emph{CVPR}, 1571--1580.

\bibitem[{Wang et~al.(2021{\natexlab{b}})Wang, Shu, Zhang, Jiang, Wang, Tian, and Wu}]{TNL2K}
Wang, X.; Shu, X.; Zhang, Z.; Jiang, B.; Wang, Y.; Tian, Y.; and Wu, F. 2021{\natexlab{b}}.
\newblock Towards More Flexible and Accurate Object Tracking with Natural Language: Algorithms and Benchmark.
\newblock In \emph{CVPR}, 13763--13773.

\bibitem[{Wang et~al.(2024)Wang, Zheng, Zhang, Cui, and Li}]{mamba-unet}
Wang, Z.; Zheng, J.-Q.; Zhang, Y.; Cui, G.; and Li, L. 2024.
\newblock Mamba-Unet: Unet-Like Pure Visual Mamba for Medical Image Segmentation.
\newblock \emph{arXiv preprint arXiv:2402.05079}.

\bibitem[{Wei et~al.(2023)Wei, Bai, Zheng, Shi, and Gong}]{artrack}
Wei, X.; Bai, Y.; Zheng, Y.; Shi, D.; and Gong, Y. 2023.
\newblock Autoregressive Visual Tracking.
\newblock In \emph{CVPR}, 9697--9706.

\bibitem[{Xie et~al.(2023)Xie, Chu, Li, Lu, and Ma}]{videotrack}
Xie, F.; Chu, L.; Li, J.; Lu, Y.; and Ma, C. 2023.
\newblock VideoTrack: Learning to Track Objects via Video Transformer.
\newblock In \emph{CVPR}, 22826--22835.

\bibitem[{Xie et~al.(2022)Xie, Wang, Wang, Cao, Yang, and Zeng}]{sbt}
Xie, F.; Wang, C.; Wang, G.; Cao, Y.; Yang, W.; and Zeng, W. 2022.
\newblock Correlation-Aware Deep Tracking.
\newblock In \emph{CVPR}, 8751--8760.

\bibitem[{Xie et~al.(2024)Xie, Zhong, Mo, Zhang, Shi, Song, and Ji}]{aqatrack}
Xie, J.; Zhong, B.; Mo, Z.; Zhang, S.; Shi, L.; Song, S.; and Ji, R. 2024.
\newblock Autoregressive Queries for Adaptive Tracking with Spatio-Temporal Transformers.
\newblock In \emph{CVPR}, 19300--19309.

\bibitem[{Xu et~al.(2020)Xu, Wang, Li, Yuan, and Yu}]{SiamFC++}
Xu, Y.; Wang, Z.; Li, Z.; Yuan, Y.; and Yu, G. 2020.
\newblock {SiamFC++:} {T}owards Robust and Accurate Visual Tracking with Target Estimation Guidelines.
\newblock In \emph{AAAI}, 12549--12556.

\bibitem[{Yan et~al.(2021{\natexlab{a}})Yan, Peng, Fu, Wang, and Lu}]{Stark}
Yan, B.; Peng, H.; Fu, J.; Wang, D.; and Lu, H. 2021{\natexlab{a}}.
\newblock Learning Spatio-Temporal Transformer for Visual Tracking.
\newblock In \emph{ICCV}, 10448--10457.

\bibitem[{Yan et~al.(2021{\natexlab{b}})Yan, Zhang, Wang, Lu, and Yang}]{Alpha-Refine}
Yan, B.; Zhang, X.; Wang, D.; Lu, H.; and Yang, X. 2021{\natexlab{b}}.
\newblock Alpha-Refine: Boosting Tracking Performance by Precise Bounding Box Estimation.
\newblock In \emph{CVPR}, 5289--5298.

\bibitem[{Ye et~al.(2022)Ye, Chang, Ma, Shan, and Chen}]{ostrack}
Ye, B.; Chang, H.; Ma, B.; Shan, S.; and Chen, X. 2022.
\newblock Joint Feature Learning and Relation Modeling for Tracking: A One-Stream Framework.
\newblock In \emph{ECCV}, 341--357.

\bibitem[{Zhang and Sennrich(2019)}]{rms}
Zhang, B.; and Sennrich, R. 2019.
\newblock Root Mean Square Layer Normalization.
\newblock In \emph{NeurIPS}, 12360--12371.

\bibitem[{Zhang and Peng(2019)}]{Deeper-wider-SiamRPN}
Zhang, Z.; and Peng, H. 2019.
\newblock Deeper and Wider Siamese Networks for Real-Time Visual Tracking.
\newblock In \emph{CVPR}, 4591--4600.

\bibitem[{Zhang et~al.(2020)Zhang, Peng, Fu, Li, and Hu}]{ocean}
Zhang, Z.; Peng, H.; Fu, J.; Li, B.; and Hu, W. 2020.
\newblock Ocean: Object-aware Anchor-free Tracking.
\newblock In \emph{ECCV}, 771--787.

\bibitem[{Zheng et~al.(2024)Zheng, Zhong, Liang, Mo, Zhang, and Li}]{odtrack}
Zheng, Y.; Zhong, B.; Liang, Q.; Mo, Z.; Zhang, S.; and Li, X. 2024.
\newblock ODTrack: Online Dense Temporal Token Learning for Visual Tracking.
\newblock In \emph{AAAI}, 7588--7596.

\bibitem[{Zhu et~al.(2023)Zhu, Lai, Chen, Wang, and Lu}]{vipt}
Zhu, J.; Lai, S.; Chen, X.; Wang, D.; and Lu, H. 2023.
\newblock Visual prompt multi-modal tracking.
\newblock In \emph{CVPR}, 9516--9526.

\bibitem[{Zhu et~al.(2024)Zhu, Liao, Zhang, Wang, Liu, and Wang}]{vim}
Zhu, L.; Liao, B.; Zhang, Q.; Wang, X.; Liu, W.; and Wang, X. 2024.
\newblock Vision Mamba: Efficient Visual Representation Learning with Bidirectional State Space Model.
\newblock \emph{arXiv preprint arXiv:2401.09417}.

\end{thebibliography}

\end{document}